\documentclass[10pt,doublespacing]{elsart}
\usepackage{latexsym,amssymb,amsmath,graphicx,epsfig,cite,bbm,float,epsf,graphics}
\usepackage{epstopdf}
\usepackage[algo2e]{algorithm2e}
\usepackage{setspace}

\journal{Signal Processing}

\renewcommand{\vec}[1]{\mbox{\boldmath${#1}$}}

\newcommand{\Real}{{\mathbbm{R}}}

\newcommand{\Rh}{\mathbbm{R}^{h}}

\newcommand{\sbt}{\left\{b(t) \right\}_{t \geq 1}}

\newcommand{\srt}{\left\{r(t) \right\}_{t \geq 1}}
\newcommand{\vhb}{\vec{\hat{b}}}

\newcommand{\vx}{\vec{x}}

\newcommand{\vr}{\vec{r}}
\newcommand{\vz}{\vec{z}}
\newcommand{\vw}{\vec{w}}
\newcommand{\vf}{\vec{f}}
\newcommand{\vu}{\vec{u}}

\newcommand{\vn}{\vec{n}}

\makeatletter

\newcommand{\Rmnum}[1]{\expandafter\@slowromancap\romannumeral #1@}
\makeatother

\def\ninept{\def\baselinestretch{1.5}}
\ninept

\begin{document}
\begin{frontmatter}

\title{Adaptive and Efficient Nonlinear Channel Equalization for Underwater Acoustic Communication}
\vspace{-0.5in}
\author[1]{Dariush Kari},
\ead{kari@ee.bilkent.edu.tr}
\author[2]{Nuri Denizcan Vanli},
\ead{denizcan@mit.edu}
\author[1]{Suleyman S. Kozat\corauthref{cor}}
\corauth[cor]{Corresponding author.}
\ead{kozat@ee.bilkent.edu.tr}
\address[1]{Department of Electrical and Electronics Engineering, Bilkent University, Ankara, Turkey, 06800}
\address[2]{Laboratory of Information and Decision Systems, Massachusetts Institute of Technology (MIT), Cambridge, MA 02139}
\vspace{-0.3in}
\begin{abstract}
We investigate underwater acoustic (UWA) channel equalization and introduce hierarchical and adaptive nonlinear channel equalization algorithms that are highly efficient and provide significantly improved bit error rate (BER) performance. Due to the high complexity of nonlinear equalizers and poor performance of linear ones, to equalize highly difficult underwater acoustic channels, we employ piecewise linear equalizers. However, in order to achieve the performance of the best piecewise linear model, we use a tree structure to hierarchically partition the space of the received signal. Furthermore, the equalization algorithm should be completely adaptive, since due to the highly non-stationary nature of the underwater medium, the optimal MSE equalizer as well as the best piecewise linear equalizer changes in time. To this end, we introduce an adaptive piecewise linear equalization algorithm that not only adapts the linear equalizer at each region but also learns the complete hierarchical structure with a computational complexity only polynomial in the number of nodes of the tree. Furthermore, our algorithm is constructed to directly minimize the final squared error without introducing any ad-hoc parameters. We demonstrate the performance of our algorithms through highly realistic experiments performed on accurately simulated underwater acoustic channels.
\end{abstract}
\end{frontmatter}
\section{Introduction}
Underwater acoustic (UWA) domain has become an important research field due to proliferation of new and exciting applications \cite{Xerri2002, mag}. However, due to poor physical link quality, high latency, constant movement of waves and chemical properties of water, the underwater acoustic channel is considered as one of the most adverse communication mediums in use today \cite{state_art, singer1,sto2}. These adverse properties of the underwater acoustic channel should be equalized by in order to provide reliable communication \cite{Patra2009,Zhang2011, shah,Iqbal2015, state_art,mag,blind_mmse,fki, itr, ch_prediction}. Furthermore, due to rapidly changing and unpredictable nature of underwater environment, constant movement of waves and transmitter-receivers, such processing should be adaptive \cite{mag, Zhang2011, Iqbal2015, fki}. However, there exist significant practical and theoretical difficulties to adaptive signal processing in underwater applications, since the signal generated in these applications show high degrees of non-stationarity, limit cycles and, in many cases, are even chaotic. Hence, the classical adaptive approaches that rely on assumed statistical models are inadequate since there is usually no or little knowledge about the statistical properties of the underlying signals or systems involved \cite{sto3,sto4,state_art}.

In this paper, in order to rectify the undesirable effects of underwater acoustic channels, we introduce a radical approach to adaptive channel equalization and seek to provide robust adaptive algorithms in an individual sequence manner \cite{dc}. Since the signals generated in this domain have high degrees of non-stationarity and uncertainty, we introduce a completely novel approach to adaptive channel equalization and aim to design adaptive methods that are mathematically guaranteed to work uniformly for all possible signals without any explicit or implicit statistical assumptions on the underlying signals or systems \cite{dc}.

Although linear equalization is the simplest equalization method, it delivers an extremely inferior performance compared to that of the optimal methods, such as Maximum A Posteriori (MAP) or Maximum Likelihood (ML) methods\cite{mimo, nergis,blind_mmse}. Nonetheless, the high complexities of the optimal methods, and also their need of the channel information\cite{shah,itr,mlse,blind_mmse,blind_rls}, make them practically infeasible for UWA channel equalization, because of the extremely large delay spread of UWA channels \cite{nergis,shah,shah2,efficient,BICM}. Hence, we seek to provide powerful nonlinear equalizers with low complexities as well as linear ones. To this end, we employ piecewise linear methods, since the simplest and most effective as well as close to the nonlinear equalizers are piecewise linear ones\cite{CTW,dc}. By using piecewise linear methods, we can retain the breadth of nonlinear equalizers, while mitigating the over-fitting problems associated with these models\cite{CTW, Hero}. As a result, piecewise linear filters are used in a vast variety of applications in signal processing and machine learning literature\cite{Hero}.

In piecewise linear equalization methods, the space of the received signal is partitioned into disjoint regions, each of which is then fitted a linear equalizer\cite{CTW,nergis}. We use the term ``linear'' to refer generally to the class of ``affine'' rather than strictly linear filters. In its most basic form, a fixed partition is used for piecewise linear equalization, i.e., both the number of regions and the region boundaries are fixed over time\cite{CTW,Hero}. To estimate the transmitted symbol with a piecewise linear model, at each specific time, exactly one of the linear equalizers is used\cite{nergis}. The linear equalizers in every region should be adaptive such that they can match the time varying channel response. However, due to the non-stationary statistics of the channel response, a fixed partition over time cannot result in a satisfactory performance. Hence, the partitioning should be adaptive as well as the linear equalizers in each region.

To this aim, we use a novel piecewise linear algorithm in which not only the linear equalizers in each region, but also the region boundaries are adaptive\cite{dc}. Therefore, the regions are effectively adapted to the channel response and follow the time variations of the best equalizer in highly time varying UWA channels. In this sense, our algorithm can achieve the performance of the best piecewise linear equalizer with the same number of regions, i.e., the linear equalizers as well as the region boundaries converge to their optimal linear solutions.

Nevertheless, due to the non-stationary channel statistics, there is no knowledge about the number of regions of the best piecewise linear equalizer, i.e., even with adaptive boundaries, the piecewise linear equalizer with a certain number of regions, does not perform well. Hence, we use a tree structure to construct a class of models, each of which has a different number of regions\cite{CTW,Gelfand}. Each of these models can be then employed to construct a piecewise linear equalizer with adaptive filters in each region and also adaptive region boundaries\cite{dc}. In \cite{Gelfand}, they choose the best model (subtree) represented by a tree over a fixed partition. However, the final estimates of all of these models should be effectively combined to achieve the performance of the best piecewise linear equalizer within this class\cite{dc}. For this, we assign a weight to each model and linearly combine the results generated by each of them. However, due to the high computational complexity resulted from running a large number of different models, we introduce a technique to combine the node estimates to produce the exactly same result. We emphasize that we directly combine the node estimates with specific weights rather than running all of these doubly exponential\cite{CTW} number of models. Furthermore, the algorithm adaptively learns the node combination weights and the region boundaries as well as the linear equalizers in each region, to achieve the performance of the best piecewise linear equalizer. Specifically, we apply a computationally efficient solution to the UWA channel equalization problem using turning boundaries trees\cite{dc}. As a result, in highly time varying UWA channels, we significantly outperform other piecewise linear equalizers constructed over a fixed partition.

In this paper, we introduce an algorithm that is shown {\em i)} to provide significantly improved BER performance over the conventional linear and piecewise linear equalization methods in realistic UWA experiments {\em ii)} to have guaranteed performance bounds without any statistical assumptions. Our algorithm not only adapts the corresponding linear equalizers in each region, but also learns the corresponding region boundaries, as well as the ``best'' linear mixture of a doubly exponential number of piecewise linear equalizers. Hence, the algorithm minimizes the final soft squared error, with a computational complexity only polynomial in the number of nodes of the tree. In our algorithm, we avoid any artificial weighting of models with highly data dependent parameters and, instead, ``directly'' minimize the squared error. Hence, the introduced approach significantly outperforms the other tree based approaches such as \cite{CTW}, as demonstrated in our simulations.

The paper is organized as follows: In section \ref{sec:prob} we describe our framework mathematically and introduce the notations. Then, in section \ref{sec:soft_partitioning} we first present an algorithm to hierarchically partition the space of the received signal. We then present an upper bound on the performance of the promised algorithm and construct the algorithm. In section \ref{sec:Simulations} we show the performance of our method using highly realistic simulations, and then conclude the paper with section \ref{sec:Conclusion}.

\section{Problem Description}\label{sec:prob}
We denote the received signal by $\srt$, $r(t) \in \mathbbm{R}$, and our aim is to determine the transmitted bits $\sbt$, $b(t) \in \{-1,1\}$. All vectors are column vectors and denoted by boldface lower case letters. For a vector $\vx$, $\vx^T$ is the ordinary transpose.

\begin{figure}
  \centering
  \includegraphics[width=0.6\textwidth]{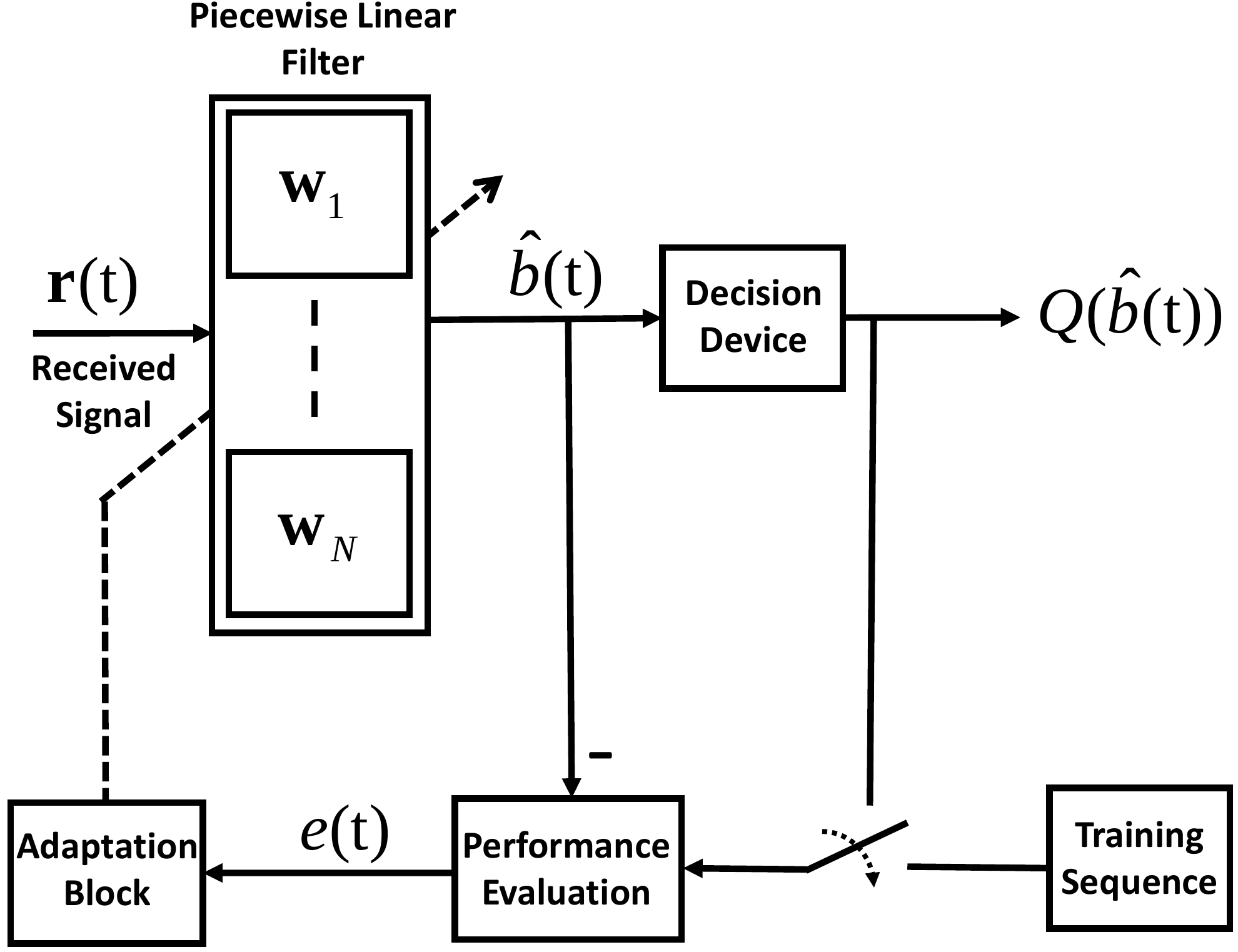}\\
  \caption{The block diagram of an adaptive piecewise linear equalizer. This equalizer consists of N different linear filters, one of which is used for each time step, based on the region (a subset of $\Rh$, where $h$ is the length of each filter) in which the received signal vector lies.}\label{fig:pla}
\end{figure}

In UWA communication, if the input signal is bandlimited, the baseband signal at the output is modeled as follows \cite{Walree}
\begin{equation}
y(t)=\sum_{p=0}^{K} g_p(t)x(t-pT_s)+\nu(t),
\end{equation}
where $y(t)$ is the channel output, $T_s$ is the sampling interval, $K$ is the minimum number beyond which the tap gains $g_p(t)$ are negligible, $\nu(t)$ indicates the ambient noise, and $g_p(t)$ is defined by
\begin{equation}
g_p(t) \triangleq \int_{-\infty}^{\infty} c(\tau,t)\ \text{sinc}(\frac{\tau-pT_s}{T_s})\ d\tau.
\end{equation}
where $c(\tau,t)$ indicates the channel response at time $t$ related to an impulse launched at time $t-\tau$, and $\tau$ is the delay time. The input signal $x(t)$ is the pulse shaped signal generated from the sequence of bits $\sbt$ transmitted every $T_s$ seconds. Note that the effects of time delay and phase deviations are usually addressed at the front-end of the receiver. Hence, we do not deal with this representation of the received signal. Instead, we assume that channel is modeled by a discrete time impulse response (i.e., a tap delay model). With a small abuse of notation, in the rest of the paper, we denote the discrete sampling times by $t$, such that the received signal can be represented as
\begin{equation}
r(t) = \sum_{k=-N_2}^{N_1} b(k) \tilde{g}(t-k) + \nu(t),
\end{equation}
where $r(t) \triangleq y(tT_s)$ is the output of the discrete channel model, $\tilde{g}(k)$ is the $k$th tap of the discrete channel impulse response, and $\nu(t)$ represents the ambient noise. We have assumed that the discrete channel can be effectively represented by $N_1$ causal and $N_2$ anti-causal taps. The input symbols $b(t)$ are transmitted every $T_s$ seconds and our aim is to estimate the transmitted bits $\sbt$ according to the channel outputs $\srt$. In this setup, a linear channel equalizer can be constructed as
\begin{equation}
\hat{b}(t) = \vw^T(t) \vr(t),
\end{equation}
where $\vr(t) \triangleq [r(t), \dots, r(t-h+1)]^T$ is the received signal vector at time $t$, $\vw(t)\triangleq [w_0(t), \ldots,w_{h-1}(t)]^T$ is the linear equalizer at time $t$, and $h$ is the equalizer length. The tap weights $\vw(t)$ can be updated using any adaptive filtering algorithm such as the least mean squares (LMS) or the recursive least squares (RLS) algorithms \cite{sayed_book} in order to minimize the squared error loss function, where the soft error at time $t$ is defined as
\[
e(t) = b(t) - \hat{b}(t).
\]
However, we can get significantly better performance by using adaptive nonlinear equalizers, because such linear equalization methods usually yield unsatisfactory performance\cite{nergis}. Thus, we employ piecewise linear equalizers, which serve as the most natural and computationally efficient extension to linear equalizers\cite{Hero},  since the equalizer lengths are significantly large in UWA channels\cite{gen}. The block diagram of a sample adaptive piecewise linear equalizer is shown in the Fig. \ref{fig:pla}. In such equalizers, the space of the received signal (here, $\Rh$) is partitioned into disjoint regions, to each of which a different linear equalizer is assigned.

\begin{figure}
  \centering
  \includegraphics[width=0.4\textwidth]{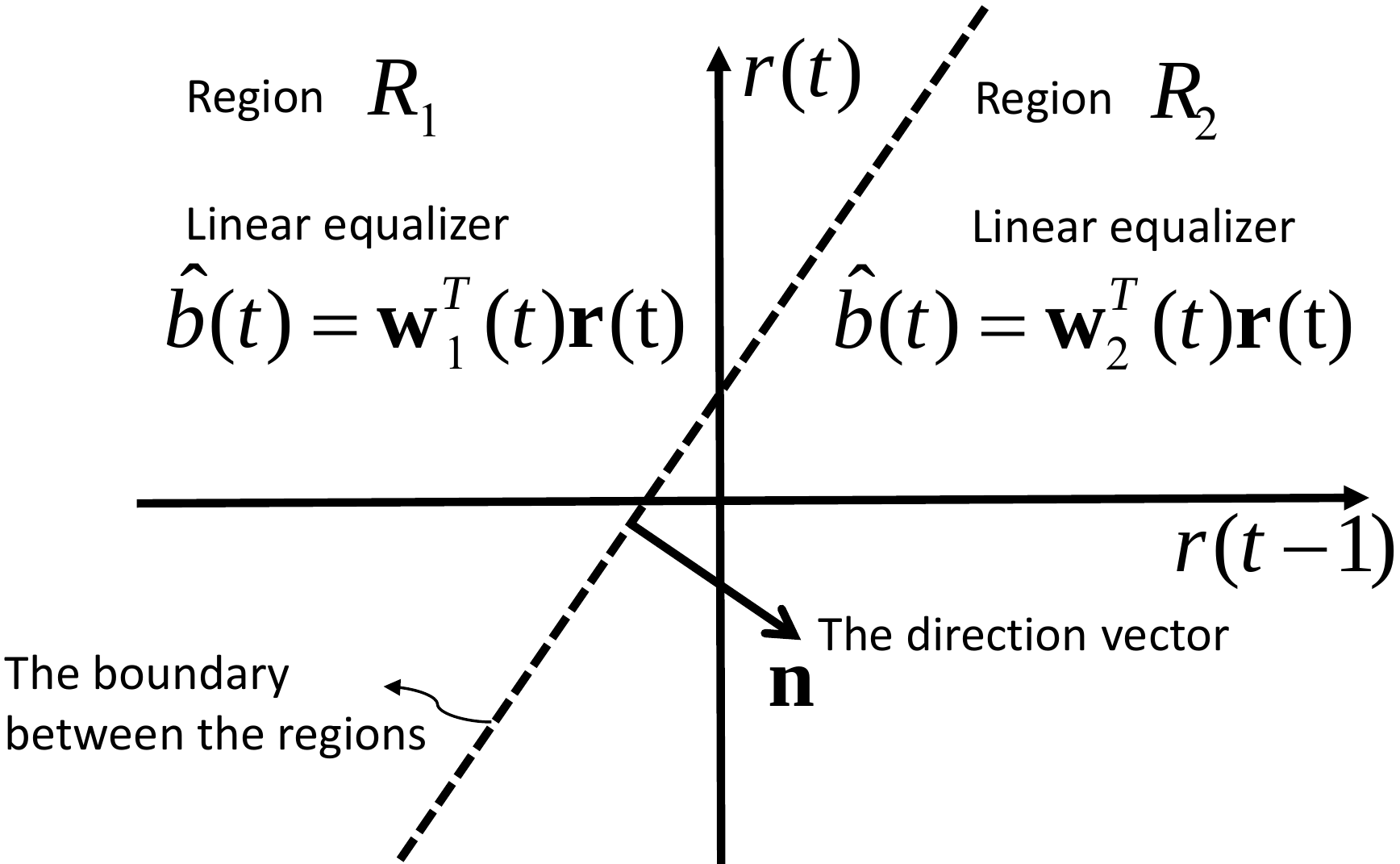}\\
  \caption{A simple two region partition of the space $\Real^2$. We use different equalizers $\vw_1$ and $\vw_2$ in regions $R_1$ and $R_2$ respectively. The direction vector $\vn$ is an orthogonal vector to the regions boundary (the hyper-plane used to separate the regions).}\label{fig:two-region}
\end{figure}

As an example, in Fig. \ref{fig:two-region}, we use the received signal $\vr(t) \triangleq [r(t),r(t-1)]^T \in \Real^2$ to estimate the transmitted bit $b(t)$. We partition the space $\Real^2$ into two regions $R_1$ and $R_2$, and use different linear equalizers $\vw_1 \in \Real^2$ and $\vw_2 \in \Real^2$ in these regions respectively. Hence the estimate $\hat{b}(t)$ is calculated as
\[
\hat{b}(t) = \begin{cases} \vw_1^{T}(t)\vr(t)+c_1(t) & \quad \textrm{if}\ \vr(t) \in R_1\\
\vw_2^{T}(t)\vr(t)+c_2(t) & \quad \textrm{if}\ \vr(t) \in R_2, \end{cases}
\]
where $c_1(t) \in \Real$ and $c_2(t) \in \Real$ are the offset terms, which can be embedded into $\vw_1$ and $\vw_2$, i.e., $\vw_j \triangleq [\vw_j^T \quad c_j]^T, j = 1,2$, and $\vr \triangleq [\vr^T \quad 1]^T$. Hence the above expression can be rewritten as
\[
\hat{b}(t) = \begin{cases} \vw_1^{T}(t)\vr(t) & \quad \textrm{if}\ \vr(t) \in R_1\\
\vw_2^{T}(t)\vr(t) & \quad \textrm{if}\ \vr(t) \in R_2. \end{cases}
\]

Because of the high complexity of the best linear  minimum mean squared error (MMSE) equalizer in each region\cite{nergis}, as well as the rapidly changing characteristics of the UWA channel, we use low complexity adaptive techniques to achieve the best linear equalizer in each region\cite{sayed_book}. Hence we update the equalizer's coefficients using least mean squares (LMS) algorithm as
\begin{align}
\vw_1(t+1) & = \vw_1(t) + \mu_1\ e(t)\ \vr(t) \quad \textrm{if}\ \vr(t) \in R_1\nonumber\\
\vw_2(t+1) & = \vw_2(t) + \mu_2\ e(t)\ \vr(t) \quad \textrm{if}\ \vr(t) \in R_2,\nonumber
\end{align}
Note that the complexity of the MMSE method is quadratic in the equalizer length\cite{nergis}, while the LMS method has a complexity only linear in the equalizer length.

To obtain a general expression, consider that we use a partition $P$ with $N$ subsets (regions) to divide the space of the received signal into disjoint regions, i.e.,
\begin{align}
& P = \{R_1, \dots, R_N\}\nonumber\\
& \Rh = \cup_{j=1}^N R_j\nonumber\\
& \hat{b}(t)=\hat{b}_j(t)=\vw_j^T(t) \vr(t) \quad \textrm{if}\ \vr(t) \in R_j,
\end{align}
which can be rewritten using indicator functions as
\begin{align}
\hat{b}(t)&=\sum_{j=1}^{N} \hat{b}_j(t)\ \textrm{id}_j(\vr(t))\nonumber\\
&=\sum_{j=1}^{N} \vw_j^T(t) \vr(t)\ \textrm{id}_j(\vr(t)),
\end{align}
where the indicator function $\textrm{id}_j(\vr(t))$ determines whether the received signal vector $\vr(t)$ lies in the region $R_j$ or not, i.e.,
\[
\textrm{id}_j(\vr(t)) = \begin{cases} 1 & \quad \textrm{if}\ \vr(t) \in R_j\\
0 & \quad \textrm{otherwise}.
\end{cases} \]

{\bf Remark 1:} Note that this algorithm can be directly applied to DFE equalizers. In this scenario, we partition the space of the extended received signal vector. To this end, we append the past decided symbols to the received signal vector as
\[
\tilde{\vr}(t) \triangleq [r(t), \dots, r(t-h+1),\bar{b}(t-1), \dots, \bar{b}(t-h_f)]^T,
\]
where $h_f$ is the length of the feedback part of the equalizer, i.e., we partition $\Real^{(h+h_f)}$. Also, $\bar{b}(t)=Q(\hat{b}(t))$ denotes the quantized estimate of the transmitted bit $b(t)$. Furthermore, corresponding to this extension in the received signal vector, we merge the feed-forward and feedback equalizers in each region to obtain an extended filter of length $h+h_f$ as
\[
\tilde{\vw}_j(t) \triangleq [\vw_j^T(t) \quad \vf_j^T(t)]^T,
\]
where $\vf_j(t)$ represents the feedback filter corresponding to the $j$th region at time $t$. Hence, the $j$th region estimate is calculated as
\[
\hat{b}_j(t) = \tilde{\vw}_j^T(t)\ \tilde{\vr}(t).
\]

In the next section, we extend these expressions to the case of an adaptive partition, both in the region boundaries and number of regions, and introduce our final algorithm.

\section{Adaptive Partitioning of The Received Signal Space}\label{sec:soft_partitioning}
\subsection{An Adaptive Piecewise Linear Equalizer with a Specific Partition}\label{sec:specific_partition}

Due to the non-stationary nature of underwater channel, a fixed partitioning over time cannot match well to the channel response, i.e., the partitioning should be adaptive. Hence we use a partition with adaptive boundaries. To this end, we use hyper-planes with adaptive direction vectors (a vector orthogonal to the hyper-plane) as boundaries. We use $\vn$ to refer to the direction vector of a hyperplane.

As an example consider a partition with two regions as depicted in Fig. \ref{fig:two-region}, hence, there is one boundary, the direction vector to which is shown by $\vn$. The indicator functions for these regions are calculated as
\begin{align}
\textrm{id}_1(\vr(t))& =\sigma(\vr(t))\nonumber\\
\textrm{id}_2(\vr(t))& =1-\sigma(\vr(t)),\nonumber
\end{align}

where

\[
\sigma(\vr(t))=\begin{cases}1 & \quad \textrm{if}\ \vr(t) \in R_1\\
0 & \quad \textrm{if}\ \vr(t) \in R_2,\end{cases}
\]

represents the hard separation of the regions. However, in order to learn the region boundaries, we use a soft separator function, which is defined as

\begin{equation}\label{eq:sigma}
\sigma(\vr) \triangleq \frac{1}{1+e^{\vr^T \vn + b}}\ ,
\end{equation}

which yields

\[
\sigma(\vr) = \begin{cases} 1 & \quad \textrm{if}\ \vr^T \vn + b \ll 0 \\
0 & \quad \textrm{if}\ \vr^T \vn + b \gg 0.  \end{cases}
\]

Although this separator function is not a hard separator function, it is a differentiable function, hence, it can be used to simply update the direction vector $\vn$ using LMS algorithm, resulting in an adaptive boundary. For simplicity, with a small abuse of notation, we redefine $\vn$ and $\vr$, as $\vn \triangleq [\vn^T \quad b]^T$ and $\vr \triangleq [\vr^T \quad 1]^T$, hence \eqref{eq:sigma} can be rewritten as
\begin{equation}\label{eq:sigma1}
\sigma(\vr) \triangleq \frac{1}{1+e^{\vr^T \vn}}\ .
\end{equation}

We use the LMS algorithm to update the direction vector $\vn$. Hence,
\begin{align}
\vn(t+1) & = \vn(t) - \frac{1}{2} \mu\ \nabla_{\vn(t)} e^2 \nonumber \\
& = \vn(t) + \mu\ e(t)\ \frac{\partial\ \hat{b}(t)}{\partial\ \vn(t)} \nonumber \\
& = \vn(t) + \mu\ e(t)\ \left(\frac{\partial\ \textrm{id}_1(\vr(t))}{\partial\ \vn(t)}\hat{b}_1(t) + \frac{\partial\ \textrm{id}_2(\vr(t))}{\partial\ \vn(t)}\hat{b}_2(t) \right) \nonumber \\
& = \vn(t) + \mu\ e(t)\ \sigma(\vr)(\sigma(\vr)-1) \left( \hat{b}_1(t) - \hat{b}_2(t) \right)\ \vr(t), \nonumber
\end{align}
since
\begin{align}
\frac {\partial \sigma(\vr)}{\partial \vn} & = \frac{-\vr\ e^{{\vr^T \vn + b}}}{(1+e^{\vr^T \vn + b})^2}\nonumber\\
& = -\vr \sigma(\vr)(1-\sigma(\vr)).
\end{align}

Since the region boundaries as well as the linear filters in each region are adaptive, if every filter converges, this equalizer can perform better than other piecewise linear equalizers with the same number of regions.

{\bf Remark:} The piecewise linear equalizers are not limited to the BPSK modulation and one can easily extend these results to higher order modulation schemes like QAM or PAM. However, for the complex valued data (e.g., in QAM modulations) the separating function should change as
\begin{equation}
	\label{eq:sigma_comp}
	\sigma(\vr) \triangleq \frac{1}{1+e^{\vr_\text{re}^T \vn_\text{re}+\vr_\text{im}^T \vn_\text{im}}}\,
\end{equation}
where the subscripts ``re'' and ``im'' denote the real and imaginary part of each vector respectively, e.g.,
\begin{align}
&\vr_\text{re} = [\text{Re}\{r(t)\},\ldots,\text{Re}\{r(t-h+1)\}]^T \nonumber \\
&\vr_\text{im} = [\text{Im}\{r(t)\},\ldots,\text{Im}\{r(t-h+1)\}]^T.
\end{align}

\subsection{The Completely Adaptive Equalizer Based on a Turning Boundaries Tree}

\begin{figure}
  \centering
  \includegraphics[width=0.7\textwidth]{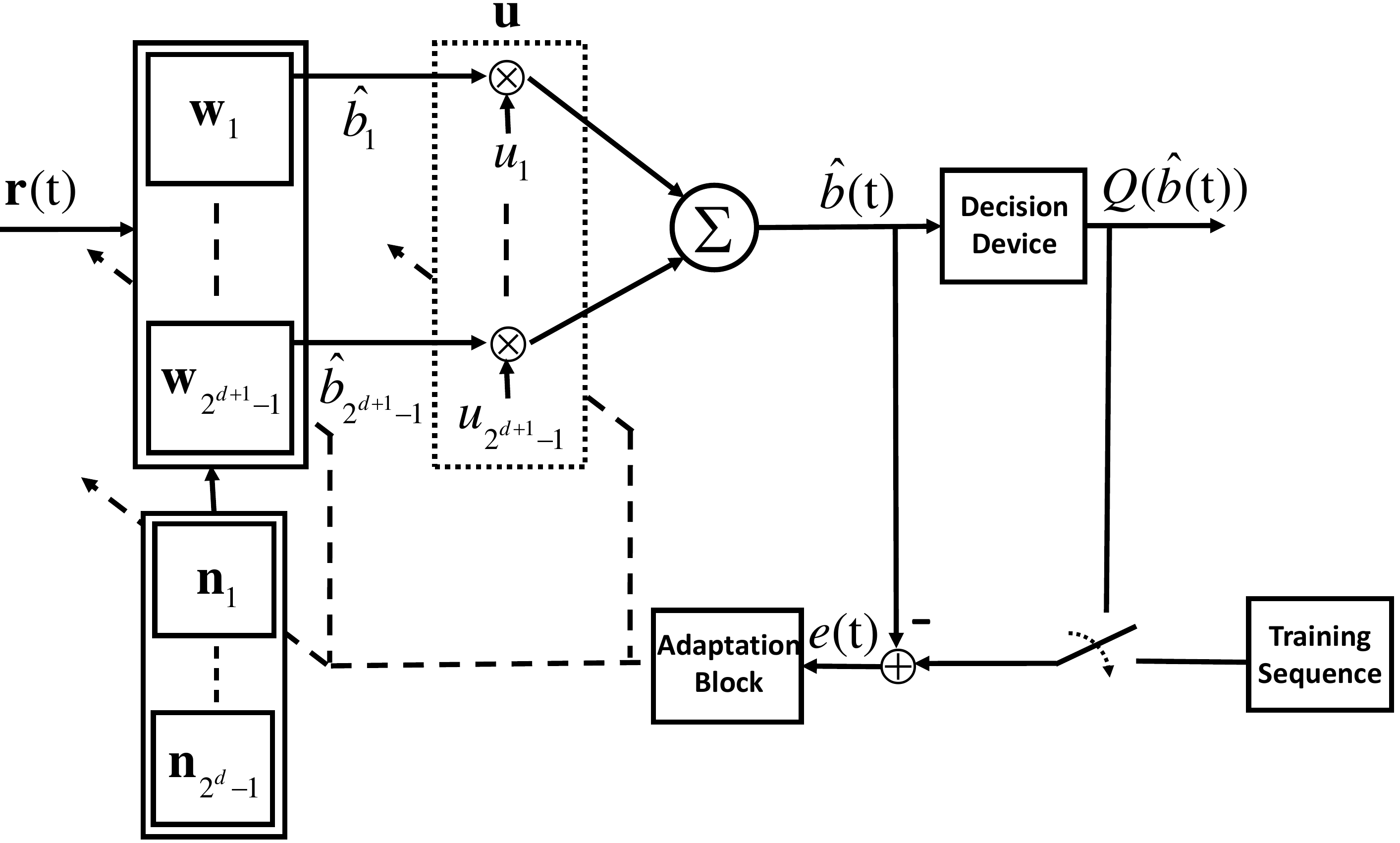}\\
  \caption{The block diagram of a turning boundaries tree (TBT) equalizer. The received signal space is partitioned using a depth $d$ tree, and corresponding to each node $i$ there is a linear filter $\vw_i$. Furthermore, the direction vectors of the separating hyper-planes, $\vn$'s, are adaptive resulting in an adaptive tree. The weight vector $\vu$, which contains the combination weights for each node's contribution, is also adaptive.}\label{fig:tbt}
\end{figure}

\begin{figure}
  \centering
  \includegraphics[width=0.7\textwidth]{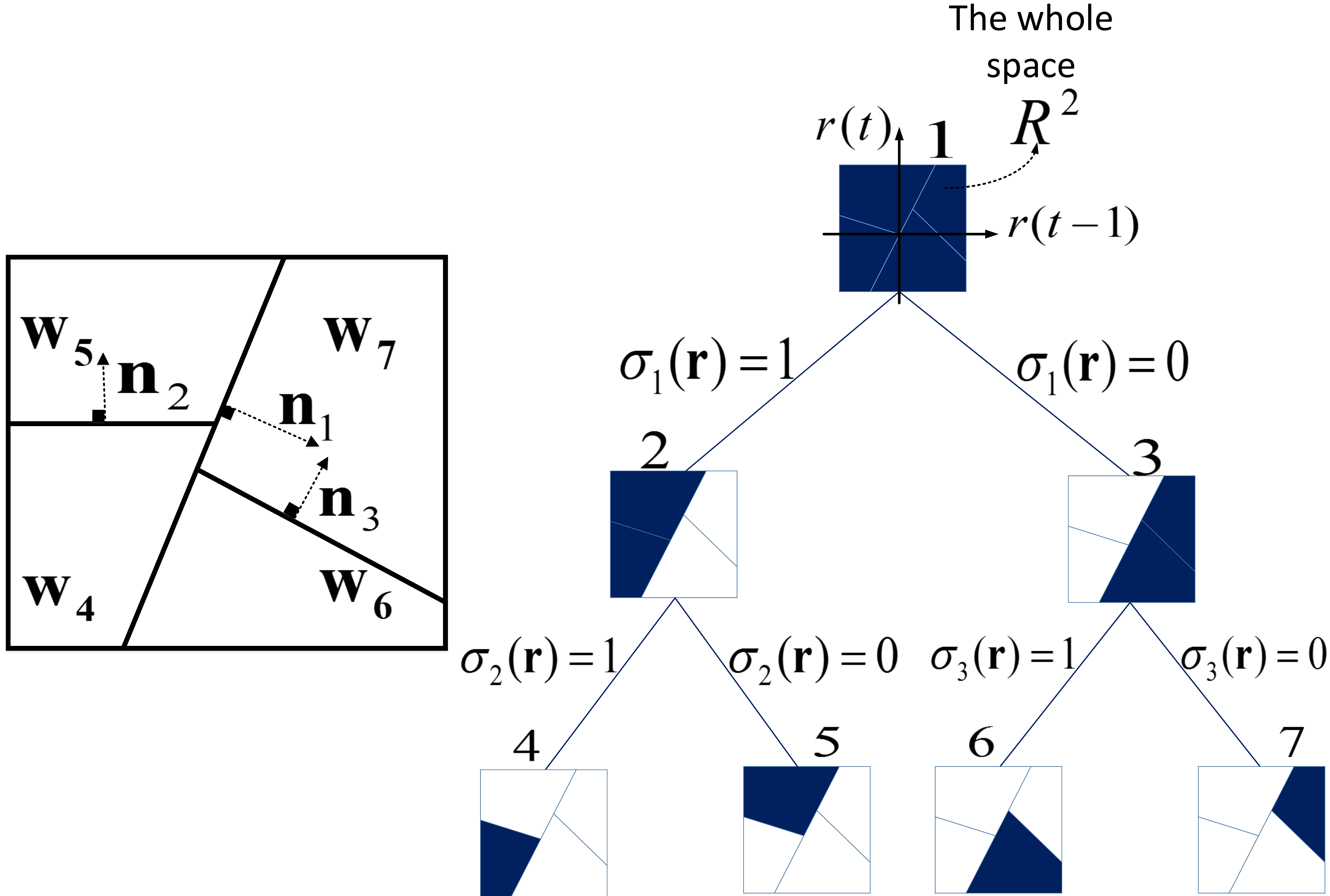}\\
  \caption{Partitioning the space $\Real^2$ using a depth-$2$ tree structure. Hyper-planes (lines) are used to divide the regions. The direction vectors are the orthogonal vectors to the hyper-planes.}\label{fig:tree}
\end{figure}

The block diagram of a sample adaptive piecewise linear equalizer with adaptive regions is shown in Fig. \ref{fig:tbt}. Given a fixed number of regions, we can achieve the best piecewise linear equalizer with the algorithm described in Section \ref{sec:specific_partition}. However, there is no a priori knowledge about the number of regions of the best piecewise linear equalizer, and the best linear equalizer will change in time, due to the highly non-stationary nature of underwater medium. In order to provide an acceptable performance with a relatively small computational complexity, we introduce a tree-based piecewise linear equalization algorithm, where we hierarchically partition the space of the received signal, i.e., $\Real^h$. Every node of the tree represents a region and is fitted a linear equalizer, as shown in Fig. \ref{fig:tree}. As shown in Fig. \ref{fig:tbt}, each node $j$ provides its own estimate $\hat{b}_j(t)$, which are then combined to generate the final estimate $\hat{b}(t)$ as
\begin{align}
\hat{b}(t)&=\sum_{j=1}^{2^{d+1}-1} u_j(t) \vw_j^{T}(t) \vr(t), \nonumber \\
& = \vu^{T}(t) \vhb(t),
\end{align}
where $\vu(t) = [u_1(t), \ldots, u_{2^{d+1}-1}(t)]^T$ is the combination weight vector, which is updated each time, and $\vhb(t) = [\hat{b}_1(t),...,\hat{b}_{2^{d+1}-1}(t)]^T$ is the vector of the node estimates.\par
\begin{figure}
  \centering
  \includegraphics[width=0.7\textwidth]{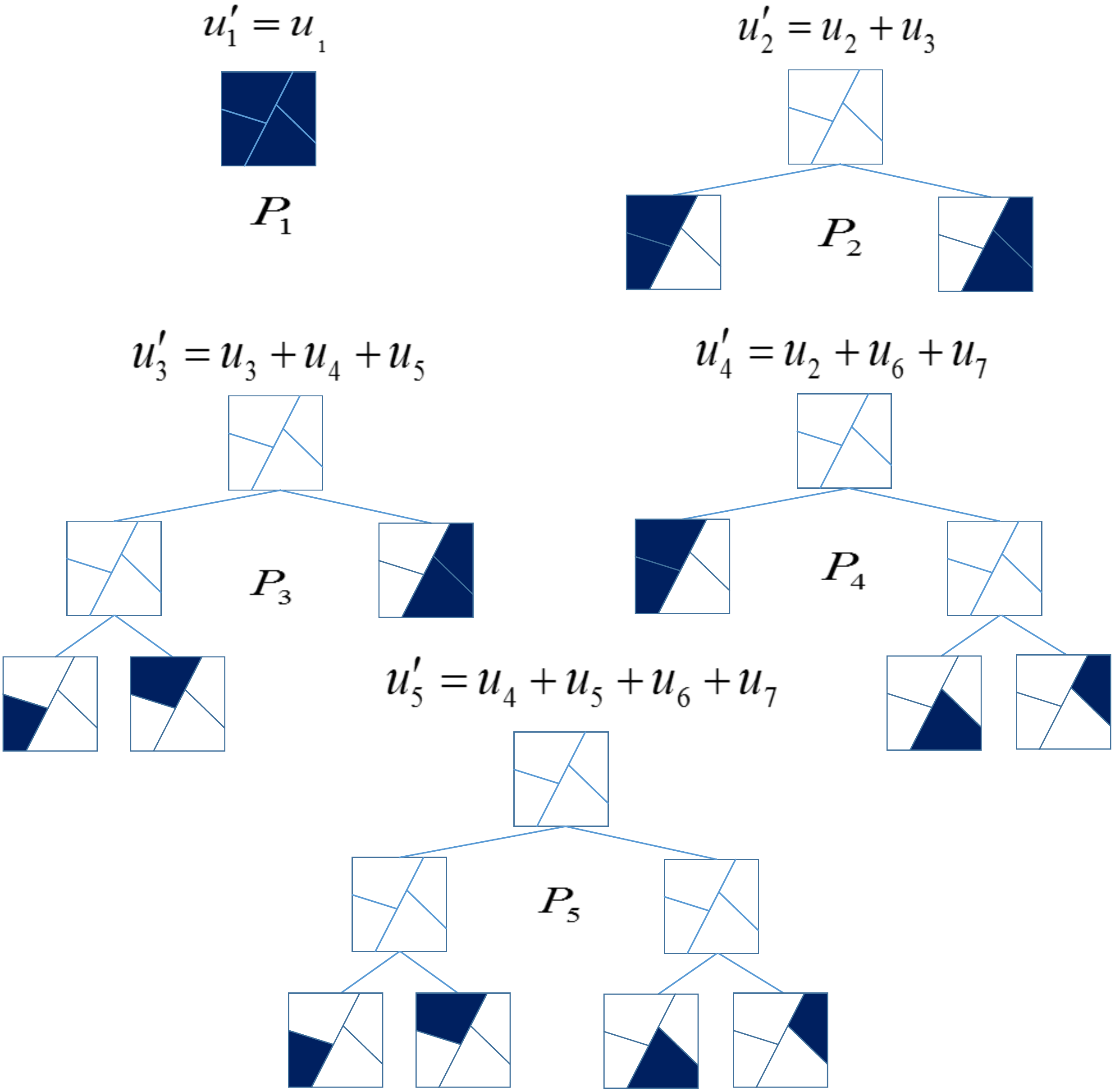}\\
  \caption{All different partitions of the received signal space that can be obtained using a depth-$2$ tree. Any of these partition can be used to construct a piecewise linear equalizer, which can be adaptively trained to minimize the squared error. These partitions are based on the separation functions shown in  Fig. \ref{fig:tree}.}\label{fig:Trees5}
\end{figure}

As depicted in Fig. \ref{fig:Trees5}, this tree introduces a number of partitions with different number of regions, each of which can be separately used as a piecewise linear equalizer\cite{dc}. Note that in our method, both the region boundaries and the channel equalizers in each region are adaptive.

To achieve the performance of the best piecewise linear equalizer, we hierarchically partition the space of the received signal. To this aim we use a tree structure in which, each node represents a region that is the union of the regions assigned to its left and right children\cite{Willems}, as shown in Fig. \ref{fig:tree}. We denote the root node by 1, and the left and right children of the node $j$ by $2j$ and $2j+1$, respectively. Obviously the root node indicates the whole space of the received signal, i.e., $\Real^h$. The estimate generated by node $j$ is calculated as
\[
\hat{b}_j(t) = \vw_j^T(t)\ \vr(t).
\]
We denote by $\alpha_d$ the number of partitioning trees with depth$\leq d$. Hence,
\[
\alpha_{d+1}=\alpha_d^2+1,
\]
which shows that there are a doubly exponential number of models embedded in a depth-$d$ tree (See Fig. \ref{fig:Trees5}), each of which can be used to construct a piecewise linear equalizer\cite{CTW}. Each of these models consists of a number of nodes. However, the number of regions (leaf nodes) in each model can be different with that of other models, as shown in Fig. \ref{fig:Trees5}, e.g., $P_2$ has two regions, while $P_5$ has 4 regions. Therefore, we implicitly run all of the piecewise linear equalizers constructed based on these partitions, and linearly combine their results to estimate the transmitted bit. We then adaptively learn the combination weights to achieve the best estimate at each time.

To clarify the framework, suppose the corresponding space of the received signal vector is two dimensional, i.e., $\vr(t) \in \mathbbm{R}^2$, and we partition this space using a depth-$2$ tree as in Fig. \ref{fig:tree}. A depth-$2$ tree is represented by three separating functions $\sigma_1(\vr(t))$, $\sigma_2(\vr(t))$ and $\sigma_3(\vr(t))$, which are defined using three hyper-planes with direction vectors $\vn_1(t)$, $\vn_2(t)$ and $\vn_3(t)$, respectively (See Fig.~\ref{fig:tree}). The left and right children of the node $j$ are $2j$ and $2j+1$ respectively, therefore, the indicator functions are defined as
\begin{align}
\textrm{id}_{1}(\vr) & = 1\nonumber\\
\textrm{id}_{2j}(\vr) & = \sigma_j(\vr)\ \times \textrm{id}_j(\vr)\nonumber\\
\textrm{id}_{2j+1}(\vr) & = (1-\sigma_j(\vr))\ \times \textrm{id}_j(\vr),\nonumber
\end{align}
where $j \leq 2^d-1$ and
\[
\sigma_j(\vr) \triangleq \frac{1}{1+e^{\vr^T \vn_j}}.
\]
Due to the tree structure, three separating hyper-planes generate four regions, each corresponding to a leaf node on the tree given in Fig. \ref{fig:tree}. The partitioning is defined in a hierarchical manner, i.e., $\vr(t)$ is first processed by $\sigma_1(\vr(t))$ and then by $\sigma_i(t)$, $i=2,3$. A complete tree defines a doubly exponential number, $O(2^{2^d})$, of models each of which can be used to partition the space of the received signal vector. As an example, a depth-$2$ tree defines 5 different partitions as shown in Fig. \ref{fig:Trees5}, where each of these subtrees is constructed using the leaves and the nodes of the original tree.

Consider the fifth model in Fig. \ref{fig:Trees5}, i.e., $P_5$, where this partition consists of $4$ disjoint regions each corresponding to a leaf node of the original complete tree in Fig. \ref{fig:tree}, labeled as $4$, $5$, $6$, and $7$. At each region, say the $4$th region, we generate the estimate $\hat{b}_{4}(t) =  \vw_{4}^T(t) \vr(t)$, where $\vw_{4}(t) \in \Real^h$ is the tap weights of the linear equalizer assigned to region $4$. Considering the hierarchical structure of the tree and having calculated the region estimates, $\hat{b}_{j}(t)$, the final estimate of $P_5$ is given by
\begin{equation}
  \hat{b}^{(5)}(t) =\sum_{j=4}^7 \textrm{id}_j(\vr(t))  \hat{b}_{j}(t),
\end{equation}
for an arbitrary selection of the separator functions $\sigma_1, \sigma_2, \sigma_3$ and for any $\vr(t)$. We emphasize that any $P_i$, $i=1,\ldots,5$ can be used in a similar fashion to construct a piecewise linear channel equalizer. Based on these model estimates, the final estimate of the transmitted bit $b(t)$ is obtained by
\begin{align}
\hat{b}(t)& =\sum_{i=1}^{\alpha_d} \hat{b}^{(i)}(t)\ u'_i(t)\nonumber\\
&  = \vhb'(t)^T\ \vu'(t),
\end{align}
where $\vhb'(t) \triangleq [\hat{b}^{(1)}(t), \dots, \hat{b}^{(\alpha_d)}(t)]^T$ and $\hat{b}^{(k)}(t)$ represents the estimate of $b(t)$ generated by the $k$th piecewise linear channel equalizer, $k=1, \dots, \alpha_d$. We use the LMS algorithm to update the weighting vector $\vu'(t)$. Note that in our method, which is given in Algorithm \ref{alg:tbt}, we linearly combine the estimates generated by all $\alpha_d$ models, using the weighting vector $\vu'(t) \triangleq [u'_1(t), \dots, u'_{\alpha_d}(t)]^T$, to estimate the transmitted bit $b(t)$, such that we can achieve the best performance on the tree.\par
Under the moderate assumptions on the cost function that $e^2(\vu'(t))$ is a $\lambda$-strong convex function \cite{Hazan} and also its gradient is upper bounded by a constant number, the following theorem provides an upper bound on the error performance of our algorithm (given in Algorithm \ref{alg:tbt}).

{\bf Theorem 1: }{\em Let $\sbt$ and $\srt$ represents arbitrary and real-valued sequences of transmitted bits and channel outputs. The algorithm for $\hat{b}(t)$ given in Algorithm \ref{alg:tbt} when applied to any sequence with an arbitrary length $L\geq1$ yields
\begin{align}\label{eq:up_bound}& E\{\sum_{t=1}^L \big{(} b(t) - \hat{b}(t) \big{)}^2\} - \min_{\vz \in \Real^{\alpha_d}}E\{ \sum_{t=1}^L \big{(} b(t) - \vz^T \vhb(t) \big{)}^2\} \leq \nonumber\\
& E\{\sum_{t=1}^L \big{(} b(t) - \hat{b}(t) \big{)}^2\} - E\{\min_{\vz \in \Real^{\alpha_d}} \sum_{t=1}^L \big{(} b(t) - \vz^T \vhb(t) \big{)}^2\} \leq  O \big{(} \log L \big{)},
\end{align}
where $\vz$ is an arbitrary constant combination weight vector, used to combine the results of all models.\\ 
{\bf Outline of the proof:}
Since we use a stochastic gradient method to update the weighting vector in Algorithm \ref{alg:tbt}, from Chapter 3 of \cite{cesa-bianchi_book} it can be straightforwardly shown that
\[
\sum_{t=1}^L \big{(} b(t) - \hat{b}(t) \big{)}^2 - \min_{\vz \in \Real^{\alpha_d}} \sum_{t=1}^L \big{(} b(t) - \vz^T \vhb(t) \big{)}^2 \leq  O \big{(} \log L \big{)},
\]
in a strong deterministic sense, which is a well known result in computational learning theory \cite{cesa-bianchi_book}. Taking the expectation of both sides of this deterministic bound yields the result in \eqref{eq:up_bound}.}

This theorem implies that the algorithm given in Algorithm \ref{alg:tbt} asymptotically achieves the performance of the optimal linear combination of the $O(2^{2^d})$ different ``adaptive'' piecewise linear equalizers, represented using a depth-$d$ tree, in the MSE sense, with a computational complexity $O(h4^d)$ (i.e., only polynomial in the number of nodes).

Regarding this theorem, for $\alpha_d \approx (1.5)^{2^d}$ different models that are embedded within a depth-$d$ tree, the introduced algorithm (given in Algorithm \ref{alg:tbt}) asymptotically achieves the same cumulative squared error as the optimal combination of the best adaptive equalizers. Moreover, note that as the data length increases and each region becomes dense enough, the linear equalizer in each region, converges to the corresponding linear MMSE equalizer in that region\cite{CTW}. In addition, since in our algorithm the tree structure is also adaptive, it can follow the data statistics effectively even when the channel is highly time varying. Therefore, our algorithm outperforms the conventional methods and asymptotically achieves the performance of the best piecewise linear equalizer.

We update the combination weights using LMS algorithm to achieve the performance of the best piecewise linear equalizer. Hence,
\begin{align*}
\vu'(t+1) & = \vu'(t)-\dfrac{1}{2} \mu \nabla_{\vu'(t)} e^2(t) \\
& = \vu'(t) + \eta\ e(t)\ \vhb(t).
\end{align*}
Note that, as depicted in Fig. \ref{fig:Trees5}, each model weight equals the sum of the weights assigned to its leaf nodes, hence we have
\[
u'_k(t) = \sum_{i \in P_k} u_i(t),
\]
which in turn results in the following node weights update algorithm
\[
u_j(t+1) = u_j(t) + \mu\ e(t)\ \hat{b}_j(t)\ \textrm{id}_j(\vr(t)),
\]
where $u_j(t)$ denotes the weight assigned to the $j$th node at time $t$.

So far we have shown how to construct a piecewise linear equalizer using separating functions and how to combine the estimates of all models to achieve the performance of the best piecewise linear equalizer. However, there are a doubly exponential number of these models, hence it is computationally prohibited to run all of these models and combine their results. In order to reduce this complexity while reaching exactly the same result, we directly combine the node estimates, i.e., instead of running all possible models, we combine the node estimates with special weights, which yields the same result. We now illustrate how to directly combine the node weights in our algorithm. The overall estimate using all models contributions is
\begin{align}
\hat{b}(t) & = \sum_{i=1}^{\alpha_d} \hat{b}^{(i)}(t)\ u'_i(t)\nonumber\\
 & = \sum_{i=1}^{\alpha_d} \hat{b}^{(i)}(t) \left(\sum_{j \in P_i} u_j(t) \right)\nonumber\\
& = \sum_{i=1}^{\alpha_d} \left(\sum_{k \in P_i} \textrm{id}_k(\vr(t))\ \hat{b}_{k}(t) \right)\left(\sum_{j \in P_i} u_j(t) \right)\nonumber\\
& = \sum_{i=1}^{\alpha_d} \left(\sum_{j,k \in P_i} \textrm{id}_k(\vr(t))\ \hat{b}_{k}(t) u_j(t) \right),
\end{align}
where $j$ and $k$ indicate two arbitrary nodes. For each node $k$, we define $z_k(t) \triangleq \textrm{id}_k(\vr(t))\ \hat{b}_{k}(t)$. Hence we have
\[
\hat{b}(t)= \sum_{i=1}^{\alpha_d} \left(\sum_{j,k \in P_i} z_k(t) u_j(t) \right).
\]
Consider that $\Gamma = \{ \Gamma_1, \dots, \Gamma_{\theta_d(d_k)} \}$ is the family of models (subtrees) in all of which the node $k$ is a leaf node,where $\theta_d(d_k)$ denotes the number of such models. Therefore the final estimate of our algorithm can be rewritten as:
\[
\hat{b}(t) = \sum_{k=1}^{2^{d+1}-1} z_k(t) \left[ \sum_{j \in \Gamma_1} u_j(t) + \dots + \sum_{j \in \Gamma_{\theta_d(d_k)}} u_j(t) \right].
\]
We denote by $\rho(j_0,k)$ the number of models in all of which the nodes $j_0$ and $k$ appear as the leaf nodes simultaneously. The weight of each node $j_0$ (i.e., $u_{j_0}$) appears in the above expression exactly $\rho(j_0,k)$ times, which yields the following expression for the final estimate
\[
\hat{b}(t) = \sum_{k=1}^{2^{d+1}-1} z_k(t)\ \beta_k(t),
\]
where
\[
\beta_k(t) \triangleq \sum_{j_0=1}^{2^{d+1}-1} u_{j_0}(t) \rho(j_0,k).
\]

We now illustrate how to calculate $\rho(j,k)$ in a depth-$d$ tree. We use $\theta_d(d_j)$ to denote the number of models extracted from a depth-$d$ tree, in all of which $j$ is a leaf node. It can be shown that
\[
\theta_d(d_j)= \prod_{l=1}^{d_j} \alpha_{d-l},
\]
where $d_j = \lfloor \log_2(j) \rfloor$ denotes the depth of the $j$th node \cite{dc}. To calculate $\rho(j,k)$ we first note that $\rho(j,k)=\rho(k,j)$ and $\rho(j,j) = \theta_d(j)$. Therefore we obtain
\[
\rho(j,k)= \begin{cases} \theta_d(j) & \quad \textrm{if}\ j=k \nonumber \\
\frac{\theta_{d-l-1}(d_k-l-1)}{\alpha_{d-l-1}}\theta_d(d_j) & \quad \textrm{if}\ j \neq k, \end{cases}
\]
where, $l$ represents the depth of the nearest common ancestor of the nodes $j$ and $k$ in the tree, i.e., an ancestor of both nodes $j$ and $k$, none of the children of that is a common ancestor of $j$ and $k$. This parameter can be calculated using the following algorithm. Assume that, without loss of generality, $j \leq k$. Obviously if $j$ is an ancestor of $k$, it is also the nearest common ancestor, i.e., $l=d_j$. However, if $j$ is not an ancestor of $k$, we define $j' \triangleq 2^{d_k-d_j}j$, which is a grandchild of the node $j$. Hence, the nearest common ancestor of $j'$ and $k$ is that of $j$ and $k$. The following procedure computes the parameter $l$.

\begin{singlespace}
\begin{algorithm2e}[H]
$l=0$\;
$\delta = d_k$\;
\While {($l \leq d_k$)}{
$\delta = \delta - l$\;
\eIf {$(j',k \leq 2^{\delta-1}+2^{\delta}\ or\ j',k \geq 2^{\delta-1}+2^{\delta})$}{
$l = l + 1$\;}{
stop\;
}
}
\end{algorithm2e}
\end{singlespace}

In order to update the region boundaries, we update their direction vectors as follows
\begin{equation}\label{eq:inner_node_update}
  \vn_j(t+1) = \vn_j(t) - \frac{1}{2} \mu \nabla_{\vn_j(t)} e^2(t),
\end{equation}
where $\nabla_{\vn_j(t)} e^2(t)$ is the derivative of $e^2(t)$ with respect to $\vn_j(t)$. Since $e(t)=b(t)-\hat{b}(t)$ the updating expression can be calculated as follows
\begin{align*}
\vn_j(t+1) & = \vn_j(t) - \frac{1}{2} \mu \nabla_{\vn_j(t)} e^2(t)\\
& = \vn_j(t) + \mu\ e(t)\ \frac{\partial\ \hat{b}(t)}{\partial\ \vn_j(t)}\\
& = \vn_j(t) + \mu\ e(t)\ \sum_{k=1}^{2^{d+1}-1} \frac{\partial\ \hat{b}(t)}{\partial\ z_k(t)}\frac{\partial\ z_k(t)}{\partial\ \vn_j(t)}\\
& = \vn_j(t) + \mu\ e(t)\ \sum_{k=1}^{2^{d+1}-1} \beta_k(t)\ \hat{b}_k(t)\frac{\partial\ \textrm{id}_k(\vr(t))}{\partial\ \vn_j(t)}\\
& = \vn_j(t) + \mu\ e(t)\ \sum_{k=1}^{2^{d+1}-1} \beta_k(t)\ \hat{b}_k(t)\frac{\partial\ \textrm{id}_k(\vr(t))}{\partial\ \sigma_j(\vr(t))}\frac{\partial\ \sigma_j(\vr(t))}{\partial\ \vn_j(t)}\\
& = \vn_j(t) + \mu\ e(t)\ \frac{\partial\ \sigma_j(\vr(t))}{\partial\ \vn_j(t)} \sum_{k=1}^{2^{d+1}-1} \beta_k(t)\ \hat{b}_k(t)\frac{\partial\ \textrm{id}_k(\vr(t))}{\partial\ \sigma_j(\vr(t))}.
\end{align*}
However note that not all of the $\textrm{id}_k(\vr(t))$ functions involve $\sigma_j(\vr(t))$, i.e., only the nodes of the subtree with the root node $j$ are included. Hence,
\\
\begin{align}
\sum_{k=1}^{2^{d+1}-1} \beta_k(t)\ \hat{b}_k(t)\frac{\partial\ \textrm{id}_k(\vr(t))}{\partial\ \sigma_j(\vr(t))} & = \sum_{m=1}^{d-d_j} \sum_{s=0}^{2^{m+1}-1}\beta_{2^m j+s}(t)\ \hat{b}_{2^m j+s}(t)\frac{\partial\ \textrm{id}_{2^m j+s}(\vr(t))}{\partial\ \sigma_j(\vr(t))}\nonumber\\
& = \sum_{m=0}^{d-d_j-1} \Big( \sum_{s=0}^{2^m-1}\beta_{2^{m+1}j+s}(t)\ \hat{b}_{2^{m+1}j+s}(t)\frac{\textrm{id}_{2^{m+1}j+s}(\vr(t))}{\sigma_j(\vr(t))}\nonumber\\
& \quad \quad \quad - \sum_{s=2^m}^{2^{m+1}-1}\beta_{2^{m+1}j+s}(t)\ \hat{b}_{2^{m+1}j+s}(t)\frac{\textrm{id}_{2^{m+1}j+s}(\vr(t))}{\sigma_j(\vr(t))} \Big).\nonumber
\end{align}
\\
We have presented the algorithm \ref{alg:tbt} for a ``turning boundaries tree'' equalizer, which is completely adaptive to the channel response. Especially in our algorithm both the number of regions and the region boundaries as well as the linear equalizers in each region are adaptive. We emphasize that the learning rates and initial values of all filters can be different.

\vspace{+1 cm}
\emph{3.3 \hspace{+0.3 cm}Complexity}
\vspace{+1 cm}
\\
Consider that we use a depth-$d$ tree to partition the space of the received signal, $\Real^h$. First note that each node estimate needs $h$ computation. Since we update all the linear filters corresponding to each region at each specific time, it generates a computational complexity of $O(h(2^{d+1}-1))$. Also, updating the separator functions results in a computational complexity of $O(h(2^d-1))$. Moreover, note that we compute the cross-correlation of every node estimate and every node weight, which results in the complexity of $O(hN_d^2)=O(h4^d)$. Hence our algorithm has the complexity $O(h4^d)$ which is only polynomial in the number of the tree nodes.
\vspace{+1 cm}
\\
From the construction of this algorithm, it can be straightforwardly shown that the algorithm is completely adaptive such that it converges to the optimal linear filters in every region and optimal partition. Therefore, the proposed equalizer achieves the performance of the best linear combination of all possible piecewise linear equalizers embedded in a depth-$d$ tree, with a complexity only polynomial in the number of tree nodes.

\begin{singlespace}
\begin{algorithm2e}[H]
\caption{Turning Boundaries Tree (TBT) Equalizer}\label{alg:tbt}
Compute $\rho(j,k)$ for all pairs $\{j,k\}$ of nodes\;
\For{$t=1\ to\ L$}{
$\vr=[r(t), \dots, r(t-h+1)]^T$\;
\For{$k = 1\ to\ 2^d-1$}{
$\sigma_k=\frac{1}{1+e^{\vr^T \vn_k}}$\;
}
$id_1 = 1$\;
\For{$k = 1\ to\ 2^d-1$}{
$id_{2k}= id_k \sigma_k$\;
$id_{2k+1}= id_k (1-\sigma_k)$\;
}
$\hat{b} = 0$\;
\For{$k = 1\ to\ 2^{d+1}-1$}{
$\hat{b}_k = \vw_k^T \vr$\;
$z_k = \hat{b}_k id_k$\;
$\beta_k = 0$\;
\For{$j = 1\ to\ 2^{d+1}-1$}{
$\beta_k = \beta_k + u_j\ \rho(j,k)$\;
}
$\hat{b}(t) = \hat{b}(t)+z_k \beta_k$\;
}
\eIf{train mode}{
$\bar{b}=b(t)$\;
}{
$\bar{b}=Q(\hat{b}(t))$\;}
$e = \bar{b} - \hat{b}(t)$\;
\For{$k = 1\ to\ 2^{d+1}-1$}{
$\vw_k = \vw_k + \mu_k\ e\ id_k \vr$\;
$u_k = u_k + \eta\ e\ z_k$\;
}
\For{$j = 1\ to\ 2^d-1$}{
$d_j = \lfloor \log_2(j) \rfloor$\;
\For{$m = 0\ to\ d-d_j-1$}{
\For{$p = 0\ to\ 2^{m}-1$}{
$i = 2^{m+1}j+p$\;
$S_1 = S_1 + \beta_i\ \hat{b}_i\ \frac{id_i}{\sigma_j}$\;}
\For{$p = 2^m\ to\ 2^{m+1}-1$}{
$i = 2^{m+1}j+p$\;
$S_2 = S_2 + \beta_i\ \hat{b}_i\ \frac{id_i}{\sigma_j}$\;}
$S = S + S_1 - S_2$\;
}
$\vn_j = \vn_j + \zeta_j\ e\ \sigma\ (\sigma-1)\ S\ \vr$\;
}
}
\end{algorithm2e}
\end{singlespace}

{\bf Remark:} Although we have introduced our equalization method in a single carrier framework, one can see that is directly extended to the OFDM framework as well. For this purpose, one can use a tree-based piecewise linear equalizer for each subcarrier in the OFDM modulation, which will improve the performance in highly nonstationary underwater acoustic channels. Furthermore, our method can be straightforwardly used in MIMO communications, i.e., one can embed all of the received symbols from all of the outputs in one vector, and then apply the proposed piecewise linear equalizer to them.\\
\section{Simulations}\label{sec:Simulations}
\begin{figure}[H]
  \centering
  \includegraphics[width=0.8\textwidth]{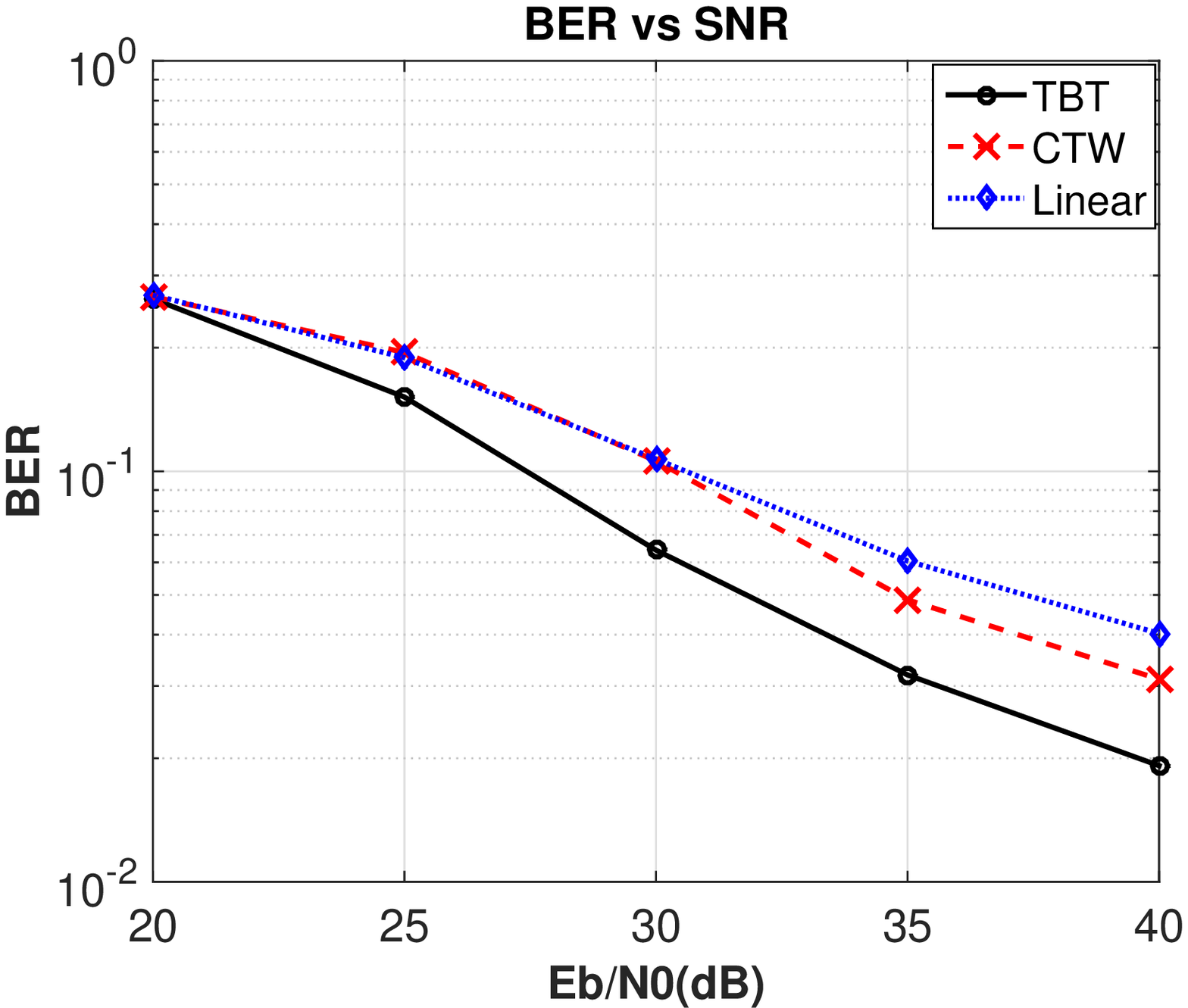}\\
  \caption{BER performances for the UWA channel response generated by \cite{gen}. The BERs for the TBT and CTW equalizers (both using depth-$2$ tree structure), and for the linear equalizer are presented.}\label{fig:ber_denizcan}
\end{figure}

\begin{figure}[H]
  \centering
  \includegraphics[width=0.8\textwidth]{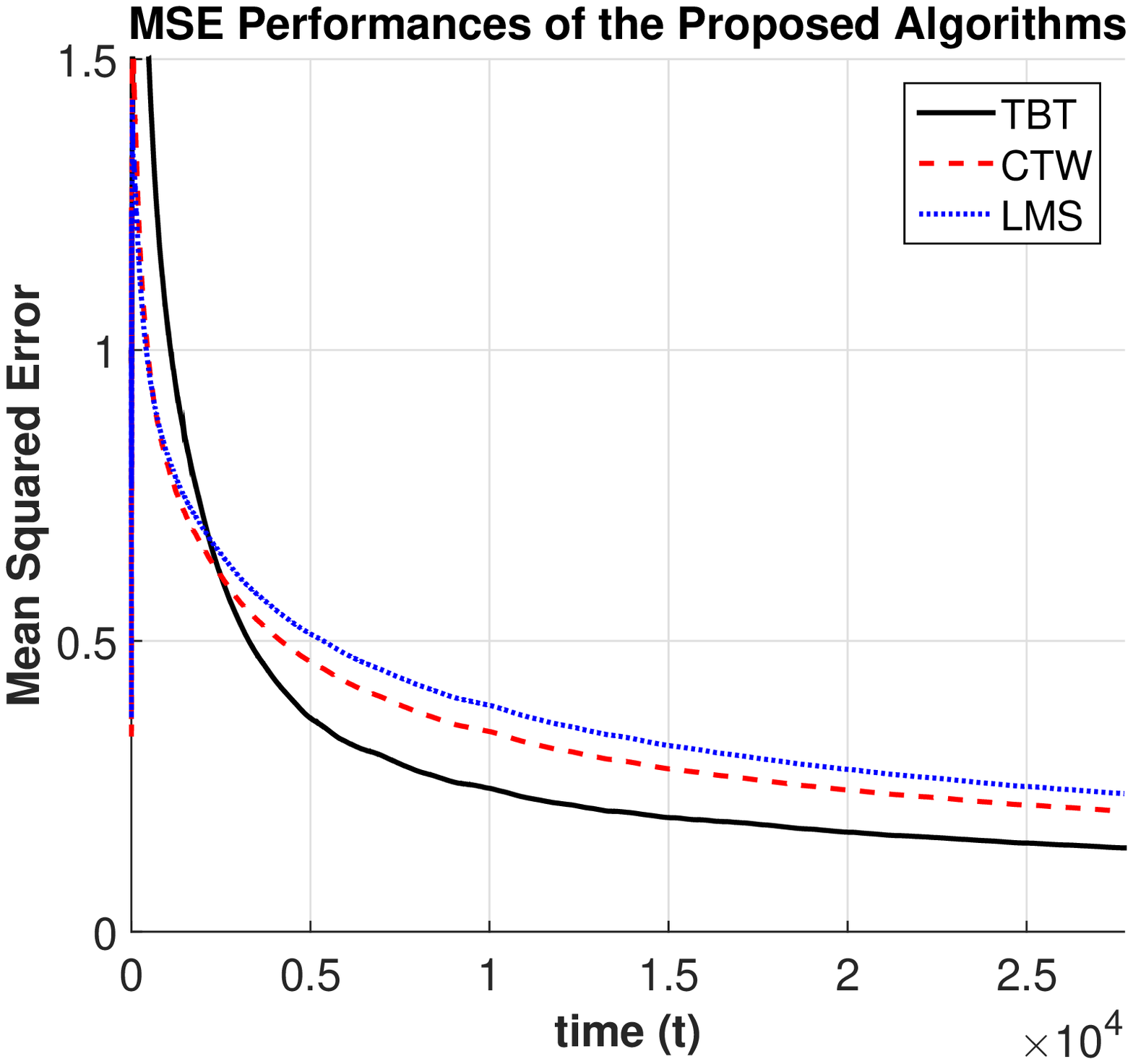}\\
  \caption{Squared error performances for the UWA channel response generated by \cite{gen} for $\mathrm{SNR}=30$dB. The time accumulated normalized squared errors for the TBT and CTW equalizers (both using depth-$2$ tree structure), and for the linear equalizer are presented.}\label{fig:mse_denizcan}
\end{figure}

In this section, we illustrate the performance of our algorithm under a highly realistic UWA channel equalization scenario. The UWA channel response is generated using the algorithm introduced in \cite{gen}, which presents highly accurate modeling of the real life UWA communication experiments as illustrated in \cite{gen}. Particularly, the surface height is set to $100$m, transmitter antenna is placed at the height of $20$m, the receiver antenna is placed at the height of $50$m, and the channel distance is $1000$m. We compare the performances of the following equalization algorithms: the Turning Boundaries Tree (TBT) equalizer of Theorem 1, the Fixed Boundaries Tree (FBT) equalizer of \cite{dc}, Finest Partition with Fixed Boundaries (FF), Finest Partition with Turning Boundaries (FT) (all having depths $d=2$), and the linear LMS equalizer. We have compared the performance of our algorithm with the context tree weighting (CTW), as shown in Fig. \ref{fig:ber_denizcan} and \ref{fig:mse_denizcan}. The Finest Partition refers to the partition consisted of all leaf nodes of the tree (the $P_5$ model in Fig. \ref{fig:Trees5}). Also, we use FBT to refer to an equalizer with fixed boundaries, which adaptively update the node weights as well as TBT algorithm. We use the LMS algorithm in the linear equalizer of each node for all algorithms, and the step sizes are set to $\mu=0.01$ for all equalizers algorithms. In all algorithms we have used length 362 equalizers.

\begin{figure}[H]
  \centering
  \includegraphics[width=0.8\textwidth]{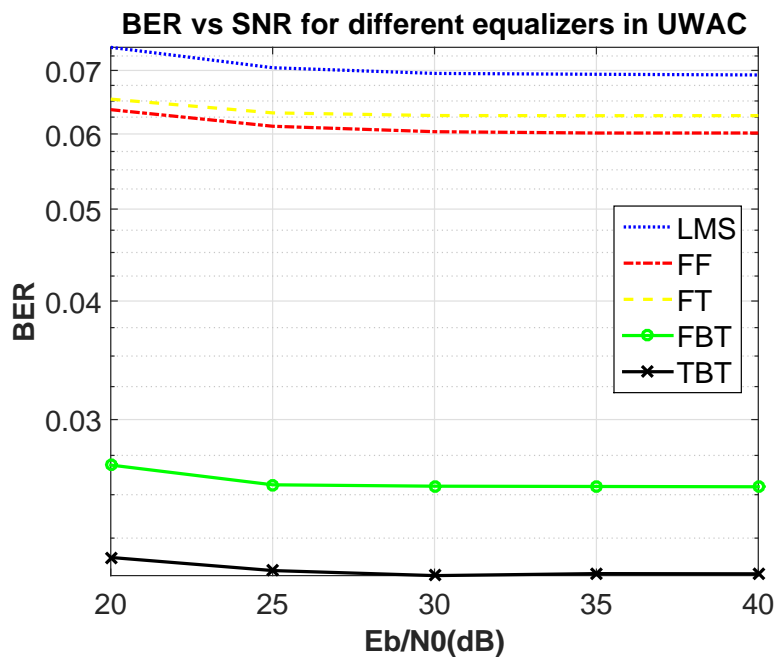}\\
  \caption{BER performances for the UWA channel response generated by \cite{gen}. The BERs for the TBT, FBT, FF and FT equalizers (all using depth-$2$ tree structure), and for the linear equalizer are presented.}\label{fig:ber}
\end{figure}

\begin{figure}[H]
  \centering
  \includegraphics[width=0.8\textwidth]{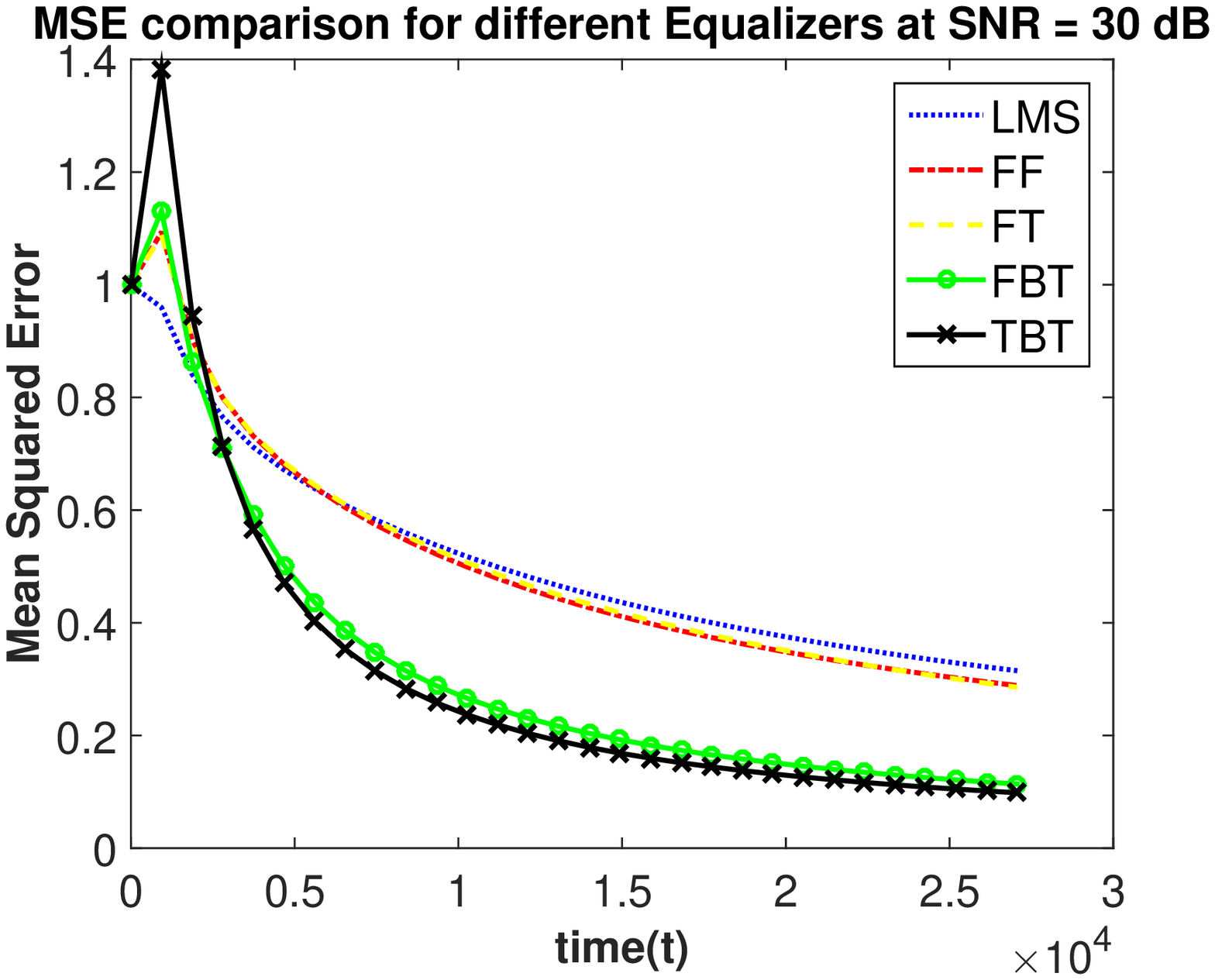}\\
  \caption{Squared error performances for the UWA channel response generated by \cite{gen} for $\mathrm{SNR}=30$dB. The time accumulated normalized squared errors for the TBT, FBT, FF, and FT equalizers (all using depth-$2$ tree structure), and for the linear equalizer are presented.}\label{fig:mse}
\end{figure}

\begin{figure}[H]
  \centering
  \includegraphics[width=0.8\textwidth]{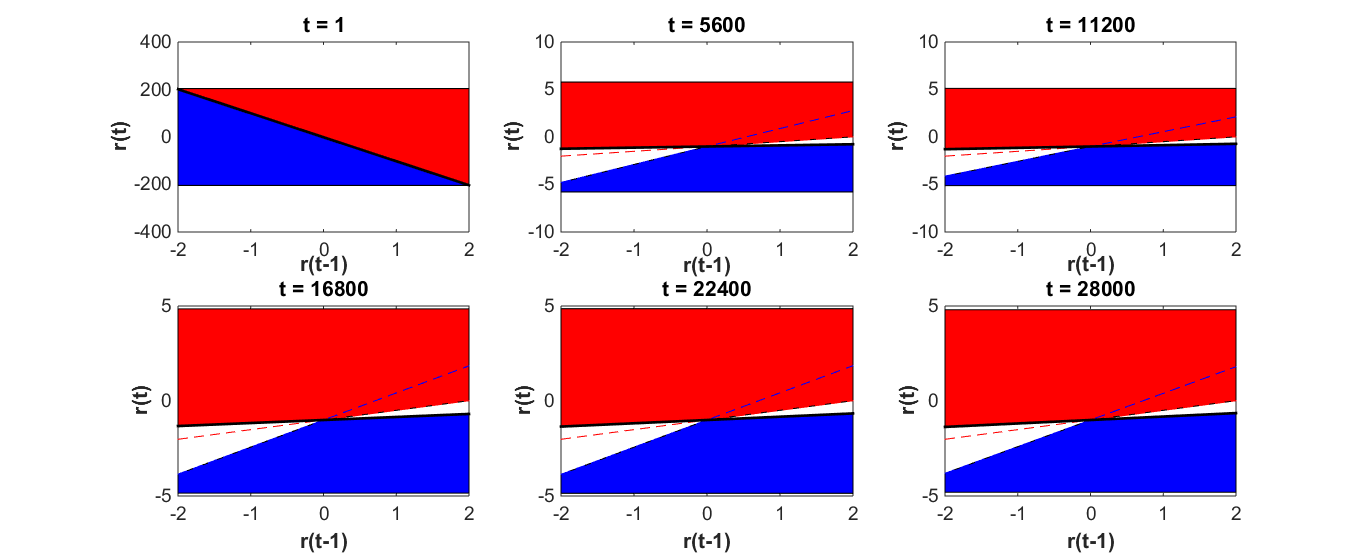}\\
  \caption{An adaptive partition using a depth 2 tree. The region boundaries are changing during the TBT algorithm converging to the optimal partition. In this experiment, $\mathrm{SNR} = 30$dB}\label{fig:regions_changing}
\end{figure}

We sent $1000$ repeated Turyn sequences \cite{turyn} (of $28$ bits) over the simulated UWA channel. Fig. \ref{fig:mse} shows the normalized time accumulated squared errors of the equalizers, when $\mathrm{SNR}=30$dB. We emphasize that the TBT equalizer significantly outperforms its competitors, where the FBT equalizer cannot provide a satisfactory result since it commits to the initial partitioning structure. Moreover, the linear equalizer yields unacceptable results due to the structural commitment to the linearity. Note that the TBT equalizer adapts its region boundaries and can successfully perform channel equalization even for a highly difficult UWA channel. The Fig. \ref{fig:ber} shows the bit error rate performance in different $\mathrm{SNR}$s for different equalizers. In the Fig. \ref{fig:regions_changing}, it is shown that the boundaries are turning during TBT algorithm, which results in an adaptive partition. The results are averaged over 10 repetitions, and show the extremely superior performance of our algorithm over other methods.

\begin{figure}[H]
  \centering
  \includegraphics[width=0.8\textwidth]{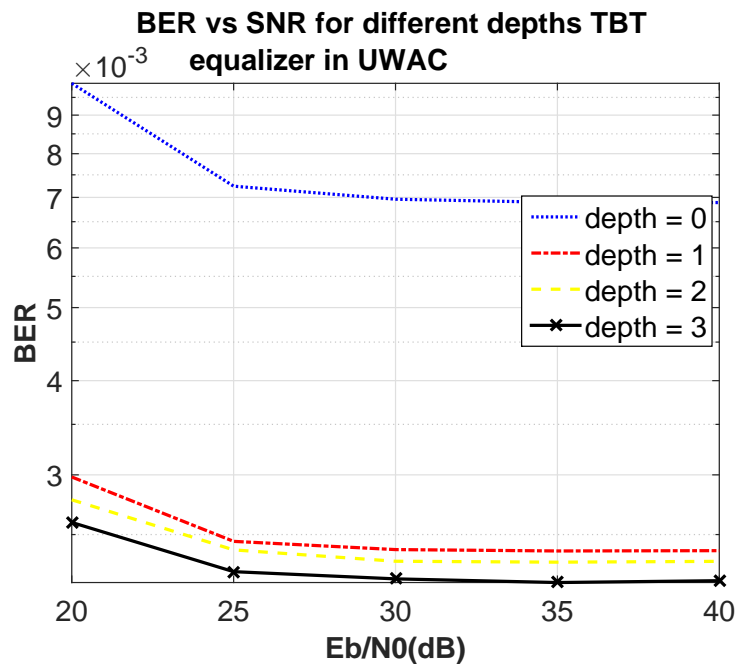}\\
  \caption{BER performances for different depths TBT equalizers. This figure shows that increasing the depth of the tree improves the BER performance.}\label{fig:ber_depth}
\end{figure}

\begin{figure}[H]
  \centering
  \includegraphics[width=0.8\textwidth]{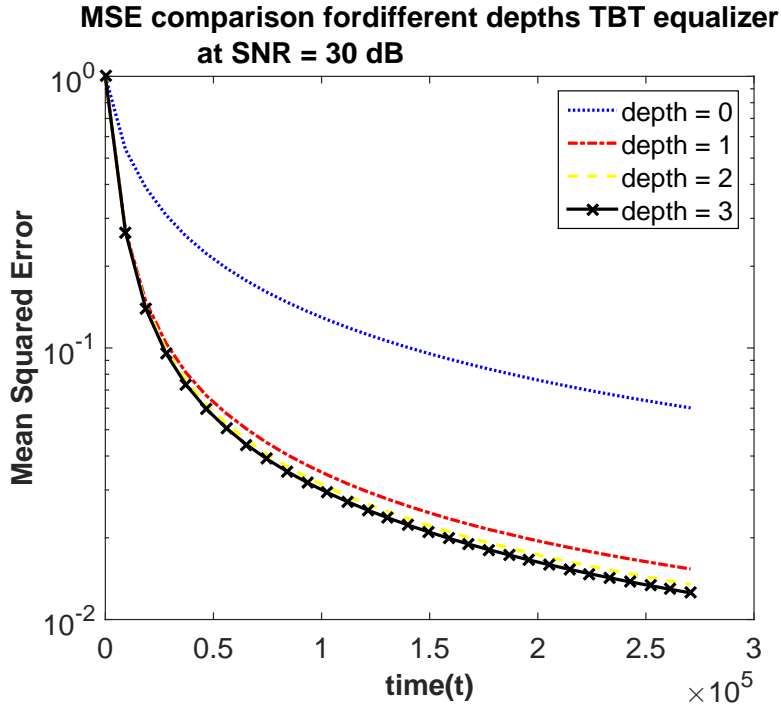}\\
  \caption{Squared error performances for different depths TBT equalizers for $\mathrm{SNR}=30$dB. This figure shows that increasing the depth of the tree improves the MSE performance. However, as the depth increases this effect diminishes.}\label{fig:mse_depth}
\end{figure}

In the second experiment we sent 10000 repeated Turyn sequence over the simulated channel, and used TBT algorithm with different depths to equalize the channel. The results, as shown in Fig. \ref{fig:ber_depth} and \ref{fig:mse_depth}, demonstrate that increasing the depth of the tree improves the performance. However, as the depth of the tree increases, the effect of the depth diminishes. This is because increasing the depth introduces finer partitions, i.e., the partitions with more regions. As the number of the regions in a partition increases, the data congestion in each region decreases, hence, the linear filters in these regions cannot fully converge to their MMSE solutions. As a result, the estimates of these regions (nodes) will be contributed to the final estimate with a much lower combination weight than other nodes, which are also present in a lower depth tree. Therefore, although increasing the depth of the tree improves the result, we cannot get a significant improvement in the performance by only increasing the depth.

\begin{figure}[H]
  \centering
  \includegraphics[width=0.8\textwidth]{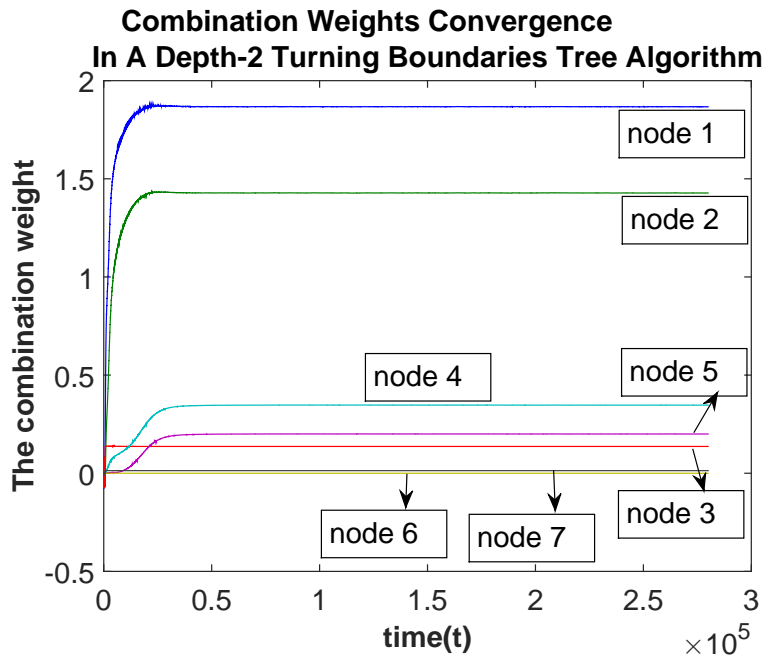}\\
  \caption{The node combination weights in TBT algorithm. In this experiment, $\mathrm{SNR} = 30$dB. Each node has a greater weight than its children, after convergence.}\label{fig:weight}
\end{figure}

Also, the node combination weights in the second experiment are shown in Fig. \ref{fig:weight}. This figure shows that node 1, the root node, has the largest weight, which means that it has the most contribution to the final estimate. Note that for an arbitrarily chosen parent node, a larger portion of the data is used to train the linear filter assigned to that node compared to its children nodes, which in turn, yields a better convergence for the parent node's filter. Hence, the contribution of the parent node to the final estimation is more than that of the children nodes. As a result the weight of each node is greater than both the weights of its left and right children, e.g., node 2 has a greater weight than node 4 and 5.

\section{Conclusion}\label{sec:Conclusion}
We study nonlinear UWA channel equalization using hierarchical structures, where we partition the received signal space using a nested tree structure and use different linear equalizers in each region. In this framework, we introduce a tree based piecewise linear equalizer that both adapts its linear equalizers in each region as well as its tree structure to best match to the underlying channel response. Our algorithm asymptotically achieves the performance of the best linear combination of a doubly exponential number of adaptive piecewise linear equalizers represented on a tree with a computational complexity only polynomial in the number of tree nodes. Since our algorithm directly minimizes the squared error and avoid using any artificial weighting coefficients, it strongly outperforms the conventional linear and piecewise linear equalizers as shown in our experiments.

\section{Acknowledgements}
This work is in part supported by Turkish Academy of Sciences, Outstanding Researcher Programme and TUBITAK, Contract No:112E161.

\bibliographystyle{elsarticle-num}
\bibliography{bibliography}

\end{document}